\documentclass[letterpaper, 10 pt, conference]{ieeeconf}  % Comment this line out if you need a4paper

\IEEEoverridecommandlockouts                              % This command is only needed if 
\usepackage{graphics} % for pdf, bitmapped graphics files
\usepackage{amsmath} % assumes amsmath package installed
\usepackage{breqn}

\usepackage{times}

\usepackage{epsfig}
\usepackage{graphicx}
\usepackage{amsmath}
\usepackage{amssymb}
\usepackage{subcaption}

\usepackage{algorithm}
\usepackage[noend]{algpseudocode}
\usepackage[algo2e]{algorithm2e}

\usepackage{multicol}
\usepackage{hyperref}
\hypersetup{colorlinks=true,urlcolor=blue,citecolor=red}

\newcommand{\norm}[1]{\left\lVert#1\right\rVert} % Command for auto adjust norm!
\newcommand{\etal}{\textit{et al}. }

\title{\LARGE \bf
Event-based Moving Object Detection and Tracking
}
\author{Anton Mitrokhin$^{1}$, Cornelia Ferm\"uller$^{2}$, Chethan Parameshwara$^{3}$, Yiannis Aloimonos$^{1}$% <-this % stops a space
\thanks{The authors are with the  Department of Computer Science$^{1}$, Institute for Advanced Computer Studies$^{2}$, and Neuroscience and Cognitive Science Program$^{3}$,  University of Maryland,  College Park, MD 20740, USA, Email: amitrokh@umd.edu, fer@umiacs.umd.edu, cmparam9@terpmail.umd.edu, yiannis@cs.umd.edu }}
\begin{document}
\maketitle
\thispagestyle{empty}
\pagestyle{empty}

%% Setting figure number to 2. Adding figure to \maketitle counts as two Fig. for some weird reason!!!
% \setcounter{figure}{1}

%%%%%%%%%%%%%%%%%%%%%%%%%%%%%%%%%%%%%%%%%%%%%%%%%%%%%%%%%%%%%%%%%%%%%%%%%%%%%%%%
\begin{abstract}
%<TODO>
%DVS is a sensor which provides unique capabilities and unique %challenges... To circumvent the challenges, most of the current %approaches are focused on reconstructing classical frames from event stream and applying well-established methods for video processing - as a result, they use only spatial information and lose a lot of data.
%We present a novel representation which enables us to utilize information about the dynamic (temporal) component of the event stream. 
Event-based vision sensors, such as the Dynamic Vision Sensor (DVS), are ideally suited for real-time motion analysis. The unique properties encompassed in the readings of such sensors provide high temporal resolution, superior sensitivity to light and low latency. These properties provide the grounds to estimate motion efficiently and reliably in the most sophisticated scenarios, but these advantages come at a price - modern event-based vision sensors have extremely low resolution, produce a lot of noise and require the development of novel algorithms to handle the asynchronous event stream.

This paper presents a new, efficient approach to object tracking with asynchronous cameras. We present a novel event stream representation which enables us to utilize information about the dynamic (temporal) component of the event stream.  The 3D geometry of the event stream is approximated with a parametric model to  motion-compensate for the  camera (without feature tracking or explicit optical flow computation), and then  moving objects that don't conform to the model are detected in an iterative process. We demonstrate our framework on the task of independent motion detection and tracking, where we use the temporal model inconsistencies to locate differently moving objects in challenging situations of very fast motion.

%The essence of the work lies in a novel representation. We model different parts of the dynamic scene using a global simple parametric model that can be efficiently derived from the dynamic (temporal) component of the event stream.
%Every event is augmented with its motion model which describes how the event stream evolves with time. This allows our algorithm without the use of  IMU or external sensors to produce motion-compensated images (effectively approximating egomotion) in extremely low-light and noisy conditions without any form of feature tracking or explicit optical flow computation. We demonstrate our framework on the task of independent motion detection and tracking, where we use the temporal model inconsistencies to locate differently moving objects in challenging situations of very fast motion.

\end{abstract}
\section*{Supplementary material}
The supplementary video materials and datasets will be made available at 
\href{http://prg.cs.umd.edu/BetterFlow.html}{http://prg.cs.umd.edu/BetterFlow.html}

The C++ implementation of the Algorithms 1 and 2 can be found here: \href{https://github.com/better-flow/better-flow}{https://github.com/better-flow/better-flow}. We provide experimental Python bindings here: \href{https://github.com/better-flow/pydvs}{https://github.com/better-flow/pydvs}

%%%%%%%%%%%%%%%%%%%%%%%%%%%%%%%%%%%%%%%%%%%%%%%%%%%%%%%%%%%%%%%%%%%%%%%%%%%%%%%%
\section{Introduction and Philosophy}
\begin{figure}[t!]
%\label{fig:intro}
\begin{center}
\resizebox{\linewidth}{!}{\begin{tabular}{cc}  
 \includegraphics[width=0.24\textwidth]{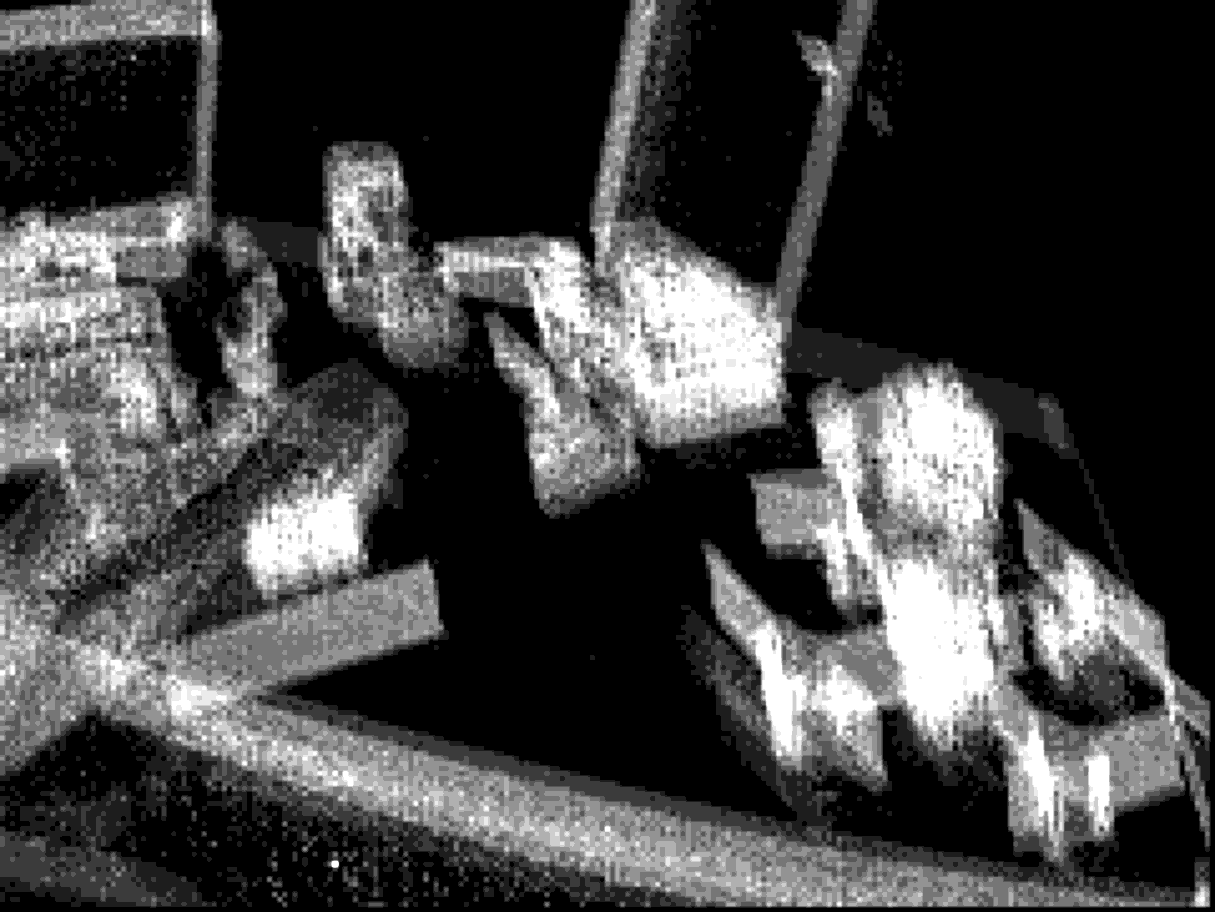}&
   \includegraphics[width=0.24\textwidth]{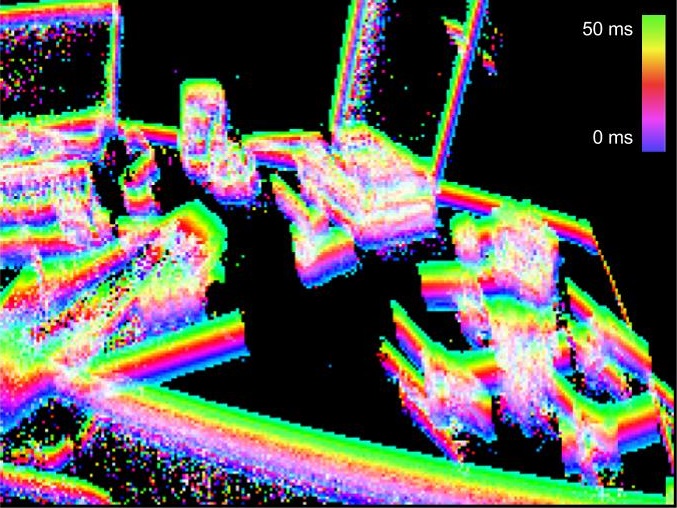}\\
   (a) & (b)\\
 \includegraphics[width=0.24\textwidth]{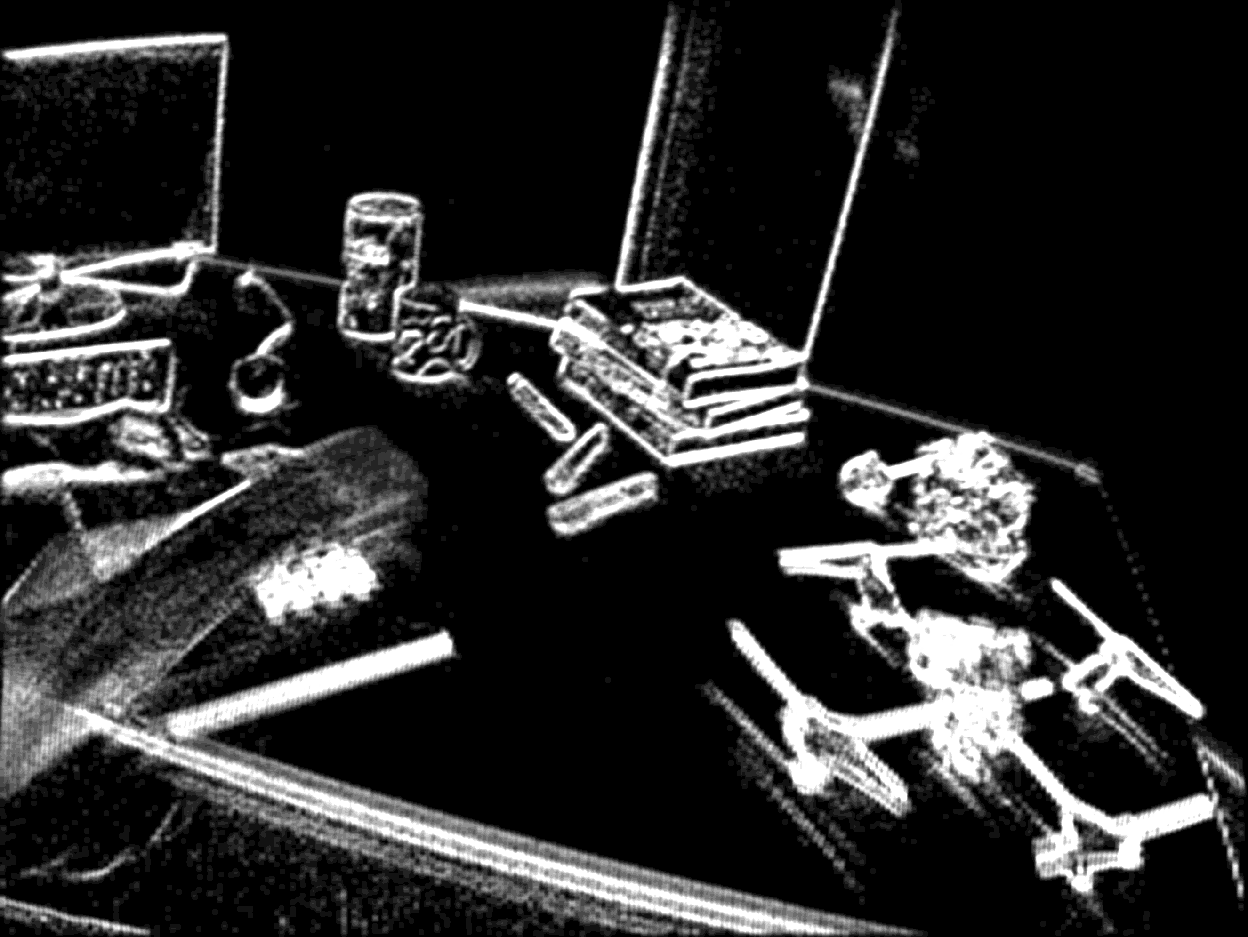}&
   \includegraphics[width=0.24\textwidth]{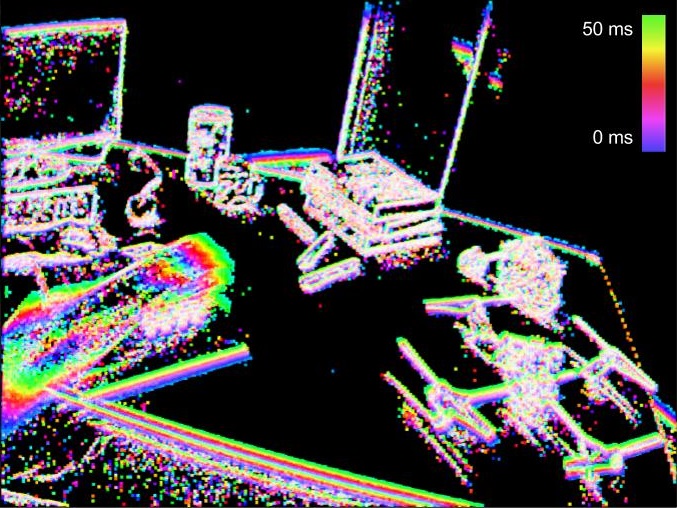}\\
    (c) & (d)
\end{tabular}}  
\vspace{-1.1\baselineskip}
\end{center}
   \caption{\small{Illustration of parametric model-fitting by alignment. The input is a cloud of events within a small time interval \(\delta t\). Visualization of event counts (a), and average time stamps (b) mapped to the image pixels after reprojection. Motion-compensated event count (c) and time image (d) acquired after the minimization stage. Colors encode the event time stamp value (with blue for oldest and green for the most recent events). Note the hand that moves independently from the camera is clearly visible on the time image, which is the basis for the subsequent detection and tracking.}}
 \label{fig:intro}
\end{figure}

The recent advancements in imaging sensor development have outpaced the development of algorithms for processing image data. Recently, the computer vision community has started to derive inspiration from the neuromorphic community whose ideas are based on biological systems to build robust and fast algorithms which run on limited computing power. These algorithms are more pervasive than ever before due to the advent of smart phones and smart cameras for a wide variety of uses from security, tracking, pursuit and mapping. 

The human fascination to understand ultra-efficient flying beings like bees, birds and flies has led  the pioneers of the field \cite{pioneers_} to conceptualize the usage of optical flow as the ultimate representation for visual motion. The availability of  optical flow makes a large variety of the aforementioned tasks simple applications of the analysis of the flow field. The computation of this superlative representation is often expensive, hence the computer vision community has come up with alternative formulations for high speed robust optical flow computation under certain constraints.  

At the same time, the robotics community follows the  approach of building 3D models for their wide spread applicability in planning algorithms and obstacle avoidance. One could accomplish any real-world task with a full 3D reconstruction of the scene from a traditional camera or other sensors. However, bio-organisms do not ``see'' the world in-terms of frames -- which is a redundant but convenient  representation used by most robotics and computer vision literature. They ``see'' the world in-terms of asynchronous changes in the scene \cite{fly_1, fly_2}. This gives unparalleled advantage in-terms of temporal resolution, low latency, and low bandwidth motion signals, potentially opening new avenues for exciting research and applications. The availability of sensors which capture events has attracted the research community at large and is gaining momentum at a rapid pace.

Most challenging problems are encountered during scenarios requiring the processing of very fast motion with real-time control of a system. One such scenario is encountered in autonomous navigation. Although computer vision and robotics communities have put forward a solid mathematical framework and have developed many practical solutions, these solutions are currently not sufficient to deal with scenes with very high speed motion, high dynamic range, and changing lighting conditions. These are the scenarios where the event based frameworks excel. 

Based on the philosophy of active and purposive perception, in this paper, we focus on the problem of multiple independently moving object segmentation and tracking from a moving event camera. However, a system to detect and track moving objects requires robust  estimates of its own motion formally known as the ego-motion estimation. We use the purposive formulation of the ego-motion estimation problem -- image stabilization. Because, we do not utilize images, we formulate a new time-image representation on which the stabilization is performed. Instead of locally computing image motion at every event, we globally obtain an estimate of the system's ego-motion directly from the event stream, and detect and track objects based on the inconsistencies in the motion field. This global model achieves high fidelity and performs well with low-contrast edges. The framework presented in the paper is highly parallelizable and can be easily ported to a GPU or an FPGA for even lower latency operation.

The contributions of this paper are:
\begin{itemize}
\item A novel event-only feature-less motion compensation pipeline.
\item A new \textit{time-image} representation based on event timestamps. This representation helps improve the robustness of  the  motion compensation.
\item A new publicly available event dataset (Extreme Event Dataset or EED) with multiple moving objects in challenging conditions (low lighting conditions and extreme light variation including flashing strobe lights).
\item A thorough quantitative and qualitative evaluation of the pipeline on the aforementioned dataset.
\item An efficient C++ implementation of the pipeline which will be released under an open-source license.
\end{itemize}

\begin{figure*}
\begin{center}
    \includegraphics[width=0.8\textwidth]{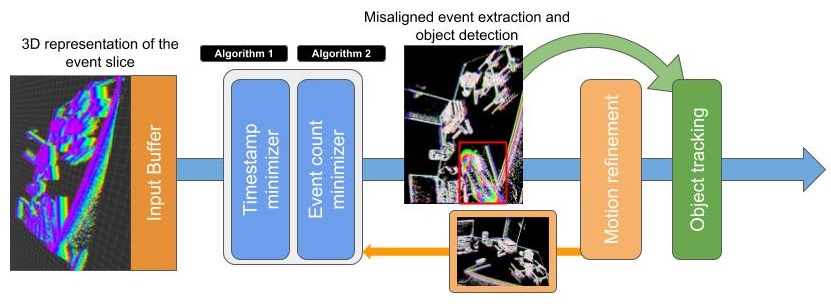}
\end{center}
\vspace{-1.0\baselineskip}
   \caption{\small{Motion compensation and independent object tracking pipeline. The event cloud motion is minimized with the error functions defined on the time-image \(\mathcal{T}\) and then, the motion is refined by maximizing the event density on the event-count image \(\mathcal{I}\). The subsets of events with high probability of misalignment (shown in red bounding box) are removed and tracked and the remaining event cloud is again motion-compensated.}}
\label{fig:pipeline}
\end{figure*}

\section{Related Work}
Over the years, several researchers have considered the problem of event-based clustering and tracking.  Litzenberger \etal \cite{litzenberger2006embedded} implemented, on an embedded system a tracking of clusters
 following circular event regions.
%simulating the mean-shift method.
Pikatkowska \etal \cite{pikatkowska2012spatiotemporal} used a Gaussian mixture model approach to track the motion of people. Linares \etal\cite{linares2015usb3} proposed an FPGA solution for noise removal and object tracking, with  clusters initialized by predefined positions. Mishra \etal\cite{mishra2017saccade} assigned events to ``spike groups'', which are clusters of events in space-time,  Lagorce \etal \cite{lagorce2015asynchronous} computed different  features and grouped them for object tracking,
and Barranco \etal \cite{barranco2018real} demonstrated a real-time mean-shift clustering algorithm.

%\begin{figure}[ht!]
%\begin{center}
%\includegraphics[width=0.8\columnwidth]{img/3D.png}
%\end{center}
%\vspace{-0.0\baselineskip}
%   \caption{\small{The motivation for the pipeline. An event stream is presented as a distorted by motion point cloud. If the background cloud (in green) is rectified, or motion-compensated, the separately moving objects (shown in blue and red) will become clearly visible after reprojecting the event cloud onto the image plane. The time-image \(\mathcal{T}\) corresponding to the current time slice is superimposed (see section \ref{sec_timestamp}) with green color denoting the most recent \textit{misaligned} events - the position of the tracked object.}}
%\label{fig:3D}
%\end{figure}

A number of works have developed event-based feature trackers (e.g. \cite{zihao2017event})
and optical flow estimators (\cite{barranco2014contour,barranco2015bio}). Based on either tracking events or features, a number of visual odometry and SLAM (simultaneous localization and mapping) approaches have been proposed. First, simplified 3D motion and scene models were considered. For example, Censi \etal\cite{censi2013low} used a known map of markers, Gallego \etal\cite{gallego2016event} considered known sets of poses and depth maps, and Kim \etal and Reinbacher \etal\cite{kim2014simultaneous,reinbacher2017real} only considered rotation. Other approaches fused events with IMU data \cite{zihao2017event}.
The  first event-based solution for  unrestricted 3D motion was presented by Kim \etal\cite{kim2016real}. Finally, a SLAM approach that combines events with images and IMU for high speed motion was recently introduced by Vidal \etal in \cite{vidal2018ultimate}.

Most closely related to our work are two studies: the work by Gallego \etal\cite{gallego2017accurate} proposes a global 3D  motion estimation approach for 3D rotation, and in \cite{gallego2018unifying} extensions of the approach are discussed. Recently in \cite{stoffregen2018simultaneous} a linear model for iterative segmentation was proposed.    Vasco \etal\cite{vasco2017independent} study  the problem of independent motion detection in a manipulation task. The expected flow field in a static scene for a given motion is learned, and then independently moving objects are obtained by tracking corners and checking for inconsistency with the expected motion. In effect, the processes of 3D motion estimation and segmentation are separated. Here, instead we consider the full problem for a system with no knowledge about its motion or the  scene. So far, no other   existing event-based approach can detect moving objects in challenging situations.

\section{Method}
Our algorithm derives inspiration from 3D point cloud processing techniques,
 such as Kinect Fusion~\cite{izadi2011kinectfusion}, 
 %and KillingFusion~\cite{slavcheva2017killingfusion}
 which use warp fields to perform a global minimization on point clouds. The algorithm performs global motion compensation of the camera by fitting a 4-parameter motion model to the cloud of events in a small time interval. These four parameters are the the x-shift, y- shift, expansion, and 2D rotation denoted as $(h_x, h_y, h_z,  \theta)$. We use two different  error functions in  the minimization in different stages of the algorithm,  which are defined in  equations \ref{eq:density} and \ref{eq:time_error_eq}. It can be shown that the 4-parameter model is a good approximation for rigid camera motion and fronto-parallel planar scene regions. The algorithm then looks for the event clusters which do not conform to the motion model and labels them as separately moving regions, while at the same time fitting the motion model to each of the detected regions. 
 %We define the error functions for the minimization with equations \ref{eq:density} and \ref{eq:time_error_eq} for different stages of the pipeline.
% We encode the displacement of points within a time interval with  4-parameters: namely the x- and y-shift, and the tangential and radial component. 

Sec.~\ref{sec_notation} describes the general notation used in this paper, and  Secs.~\ref{sec_density} and \ref{sec_timestamp}  provide an intuition for the error functions used for motion compensation. The details of the algorithm are described in Sec.~\ref{sec_main_alg} (motion compensation) and \ref{sec_track_alg} (object detection and tracking).

\section{Preliminaries}\label{sec_preliminaries}

%\subsection{The Motion Model}
%\label{se:model}
%We model the optic flow $(u, v)$ with four parameters: the shifts in $x-$ and $y-$ direction, the radial and the tangential component, which we denote as $ \bs{s} = (s_x,s_y, s_r, s_{\theta})$. The latter are computed from the divergence and the curl.
%Approximating  the optical flow with a first order Taylor expansion as
%\begin{equation}
%\begin{array}{lcl}
%u &= &u_0 + u_x x + u_y y\\
%v &=& v_0  + v_x x + v_y y,\\
%\end{array}
%\end{equation}
%where $u_0, v_0$ denote the constants and $u_x,u_y,v_x,v_y$ the partial derivatives of $u, v$ in  $x$ and $y$,
%The constant, radial and tangential components are: $u_0; v_0, u_y y - v_x x;  u_x x + v_y y$,
%and our parameters amount to
%$s_x = u_0, s_y = v_0, s_r =\arctan\frac{-v_x x}{u_y y}, s_t = ||u_x x + v_y y||$. 

%<to do> Then maybe show what these components are for a rigidly plane with inverse depth 1/Z = a*x + b*y + c as follows
%The optic flow of a rigid motion is give as:
%...

\subsection{Notation}
\label{sec_notation}
Let the input events in a temporal segment $[t_0, t_0 + \delta t]$  be represented by a 3-tuple  $C \{x, y, t\} \in \mathbb{R}^3$. Here $\left\{ x, y\right\}$ denote the spatial coordinates in the image plane and $t$ denotes the event timestamps\footnote{The data from the DVS sensor is four-dimensional, with the additional, fourth component  a binary value denoting the sign of intensity change. However, because of noise at object boundaries, we do not utilize this value here.}.  $t_0$ can be arbitrarily chosen, as the DVS data is continuous. However, the algorithm functions on a small time segment $\delta t$. 

% While the event stream from the DVS camera is continuous and we impose no restrictions on when the computation may be initiated, we define our algorithm for events within a certain small temporal segment: \(t_0\) is the beginning of the time segment and \(\Delta t\) is the width of the segment.

%  We define our input \(C\) as a set of triples \(\{x, y, t\} \in \mathbb{R}^3\) where \(x\) and \(y\) are the event spatial coordinates in the image plane  and \(t \in [t_0, t_0 + \Delta t]\) are event timestamps.\footnote{The data from the sensor is four-dimensional, with the additional, fourth component being a binary value denoting the ``direction'' of intensity change. However, because of noise at object boundaries, we do not utilize this value here.}  While the event stream from the DVS camera is continuous and we impose no restrictions on when the computation may be initiated, we define our algorithm for events within a certain small temporal segment: \(t_0\) is the beginning of the time segment and \(\Delta t\) is the width of the segment.

Let us denote the 2D displacement  that maps  events  $(x,y)$ at time $t$ to their locations $(x', y')$  at time $t_0$  by a warp field $\phi(x,y, t-t_0) :(x,y,t) \xrightarrow{} (x',y',t)$.
Our goal is to find the motion compensating warp field \(\phi : \mathbb{R}^3 \to \mathbb{R}^3\) such that the motion compensated events, when projected onto the image plane, have maximum density. Let us denote these motion compensated events as:

\begin{align}
C' = \Pi\{\phi(C)\} =\Pi\{\phi(x, y, t - t_0)\}\\
= \{x', y', 0\}\,  \quad \forall \{x, y, t\} \in C \nonumber
\label{eq:C}
\end{align}
% $\forall \{x, y, t\} \in C$

Due to the geometric properties of the events, such a warp field encodes the per-event optical flow. Here $\Pi$ is the temporal projection function projecting motion compensated events along the time axis. $\Pi$ is used to reduce the dimensionality of the data from \(\mathbb{R}^3 \to \mathbb{R}^2\) and simplify the minimization process. We represent  the data available in $C$ with  two discretized maps which encode  the temporal and intensity properties of the event stream.

\subsection{Event Count Image}\label{sec_density}
To calculate the event density $\mathcal{D}$ of $C' = \Pi\{\phi(C)\}$, we discretize the image plane into pixels of a chosen size. We use symbols \((i, j) \in \mathbb{N}^2\) to denote the integer pixel (discretization bin) coordinates, while \((x', y', t) \in \mathbb{R}^3\) represent real-valued warped event coordinates. Each projected event from \(C'\) is mapped to a certain discrete pixel, and the \textbf{total number of events mapped to the pixel is recorded as a value of that pixel}. We will henceforth refer to this data structure as \textit{event-count image} $\mathcal{I}$. Let 

\begin{equation}
    \xi_{ij} = \{\{x', y', t\} : \{x', y', 0\} \in C',\, i = x',\, j = y'\}
\label{eq:xi_trajectory}
\end{equation}

be the \textit{proposed event trajectory} - a set of warped events along the temporal axis which get projected onto the pixel \((i, j)\) after the \(\phi\) operation has been applied. Then the event-count image pixel \(\mathcal{I}_{ij}\) is defined as: 
\begin{equation}
	\mathcal{I}_{ij} = \vert\xi_{ij}\vert
\label{eq:ec_img}
\end{equation}
Here, $\vert A \vert$ is the cardinality of the set $A$.
The event density $\mathcal{D}$ is computed as:
\begin{equation}
	\mathcal{D} = \frac{\sum_{i,j} \mathcal{I}_{ij}}{\# \mathcal{I} } = \frac{\vert C'\vert }{\# \mathcal{I}}
\label{eq:density}
\end{equation}
where,  \(\# \mathcal{I}\) denotes the number of pixels with at least one event mapped on it.
Since \(\vert C' \vert\) is a constant for a  given time slice, the problem can be reformulated as the minimization of the \textit{total area} \(S = \# \mathcal{I}\) on the event-count image. A similar formulation was used in \cite{gallego2017accurate} to estimate the rotation of the camera with the acutance as an error metric, instead of the area.

%\begin{figure}[ht!]
%\begin{center}
%  \begin{minipage}[b]{1.0\linewidth}
%    \includegraphics[width=1.0\textwidth]{img/geometric.png}
%  \end{minipage}
%\end{center}
%\vspace{-1.3\baselineskip}
%   \caption{\small{Geometric representation of the event cloud cross section in a \((x, t)\) %plane}}
%\label{fig:geometric}
%end{figure}

%Fig. \ref{fig:geometric} provides a geometric interpretation of the minimization for one parameter. It shows  a cross section of the event cloud (where only x and t axises are shown for better visibility). \(PQTN\) represents a slice of an event cloud in \((x, t)\) plane and discrete events are approximated by a continuous area. We parametrize the projection area \(S\) with a single angle \(\alpha\ \in [0, \frac{\pi}{2}]\), given in respect to \(\vec{PQ}\). It is easy to see that
%\begin{equation}
%	S(\alpha) = PN + |\Delta t (\cot(\alpha) + \cot(\frac{\pi}{2} - \angle QPN + \alpha))|
%\label{eq:S}
%\end{equation}
%is a convex function which has a gradient everywhere (except for its minimum) with a minimum at \(\alpha = \frac{\pi}{2}\). Hence, a simple gradient descent can be used to find the minimum. The resulting value of \(\angle QPN\) represents the local optical flow and is a part of the compensating warp field \(\phi\).

\subsection{Time-image \texorpdfstring{$\mathcal{T}$}{T}}
\label{sec_timestamp}
Keen readers would observe that the event-count image representation $\mathcal{I}$ suffers from a subtle drawback. When the projection operation is performed, events produced by different edges (edges corresponding to different parts of the same real-world object or different real-world objects) can get projected onto the same pixel. This is a very common situation which occurs during fast motion in highly textured scenes.

% An obvious failure case of the event-count image representation is when events from different edges \textbf{explain this part, what is events from different edges?} get projected to the same pixel. This may happen during fast motion in highly textured scenes.
% In order to circumvent this, we leverage the information from the event timestamps \(t\) by proposing a novel representation which we refer to as a \textit{timestamp image}. Similar to the \textit{event-count image} described above, the timestamp image is a discretized plane with each cell containing the \textit{average} timestamp of the events mapped to this cell by the warp field \(\phi\).

To alleviate this problem, we utilize the information from the event timestamps \(t\) by proposing a novel representation which we call the \textit{time-image} $\mathcal{T}$. Similar to $\mathcal{I}$ described before, $\mathcal{T}$ is a discretized plane with each pixel containing the \textit{average} timestamp of the events mapped to it by the warp field \(\phi\).

\begin{equation}
	\mathcal{T}_{ij} = \frac{1}{\mathcal{I}_{ij}}\sum t : t \in \xi_{ij}
\label{eq:ts_img}
\end{equation}

Computing the mean of timestamps allows us to increase fidelity of our results by making use of all available DVS events. An alternative approach would be to consider only the latest timestamps \cite{zhu2018ev}, where performance would suffer in low light situations. This is because the signal to noise ratio in DVS events depends on the average illumination, the smaller the average illumination, the larger the noise. Note that the value of a time-image pixel \({\mathcal{T}}_{ij}\) (or better its deviation from the mean value) correlates with the probability that the motion was not compensated locally - this will be used later for motion detection in Sec.~\ref{sec_detection}. $\mathcal{T}$ follows the 3D structure of the event cloud, and its gradient $(G)$ provides the global metrics of the motion-compensation error which will be minimized by the algorithm:
\begin{equation}
	Error = \sum \norm{G[i, j]} = \sum (G_{x}^2[i, j] + G_{y}^2[i, j])
\label{eq:time_error_eq}
\end{equation}

Here $(G_x[i, j], G_y[i, j])$ denote the local spatial gradient of $\mathcal{T}$ along the $(x)$ or $(y)$ axes. Equation \ref{eq:time_error_eq} is a global error which takes into account local motion inconsistencies of the warped event cloud. Together with the rigid body motion assumption, equation \ref{eq:time_error_eq} can be decomposed into equations:

% As will be shown later (section \ref{sec_detection}), the value of the pixel \({\mathcal{T}}_{ij}\) correlates with the probability that the motion was not compensated locally - this will be used later for motion detection. The timestamp image follows the 3D structure of the event cloud and provides gradients for the global motion model:

% Another approach would be, as presented in \cite{latest_flow}, to consider only the latest timestamps but this would not allow the algorithm to perform well in low light \textbf{why? small explanation would be good}. As will be shown later, the value of the pixel \({T}_{ij}\) correlates with the probability that the motion was not compensated locally - this will be used later for motion detection. The timestamp image follows the 3D structure of the event cloud and provides gradients for the global motion model:

\begin{equation}
	d_{x} = \frac{\sum G_{x}[i, j]}{\# \mathcal{I}},\qquad d_{y} = \frac{\sum G_{y}[i, j]}{\# \mathcal{I}}
\label{eq:shift_eq}
\end{equation}

\begin{equation}
   	d_{z} = \frac{\sum (G_x[i, j], G_y[i, j]) \cdot (i, j)}{\# \mathcal{I}}
\label{eq:curl_eq}
\end{equation}

\begin{equation}
	d_{\theta} = \frac{\sum (G_x[i, j], G_y[i, j]) \times (i, j)}{\# \mathcal{I}}
\label{eq:div_eq}
\end{equation}
Equations \ref{eq:shift_eq}, \ref{eq:curl_eq} and \ref{eq:div_eq} will be used  to provide gradients for the motion compensation algorithm in Sec. \ref{sec_models} - each of them correspond to the error for one of the model parameters: $(h_x, h_y, h_z, h_{\theta})$, which represent shift in the image plane, expansion and 2D rotation.

\subsection{Minimization Constraints}
\label{sec_models}
The local gradients of $\mathcal{T}$ and the event density $\mathcal{D}$ both quantify the error in event cloud motion compensation. %However, the problem of optimizing for a warp field \(\phi\), which minimizes the local error at every point of the timestamp and event count images is hard. To simplify this optimization problem, 
We model the global warp field \(\phi^G(x, y, t)\) with a 4 parameter global motion model \(\mathcal{M}^G = \{h_x, h_y, h_z, \theta\}\) to describe the distortion induced in the event cloud by the camera motion. The resulting coordinate transformation amounts to: %Eq.~\ref{eq:transform}.

% Still, the problem of computing a warp field \(\phi\) to minimize the local error at every point of the timestamp and event count images is rather hard.

% We model the global warp field \(\phi^G(x, y, t)\) with a 4 parameter global motion model \(\mathcal{M^G} = \{h_x, h_y, h_z, \theta\}\) which has shown to describe the distortion induced in the event cloud by the camera motion. The resulting coordinate transformation is shown in equation (\ref{eq:transform}).

\begin{equation}
\resizebox{0.9\columnwidth}{!}{$
    \begin{bmatrix}x' \\ y' \end{bmatrix} = \begin{bmatrix}x \\ y \end{bmatrix} - t * \Bigg[\begin{bmatrix}h_x \\ h_y \end{bmatrix} + (h_z + 1) * \begin{vmatrix}\cos{\theta} & -\sin{\theta} \\ \sin{\theta} & \cos{\theta} \end{vmatrix} \cdot \begin{bmatrix}x \\ y \end{bmatrix} - \begin{bmatrix}x \\ y \end{bmatrix} \Bigg]
$}
\label{eq:transform}
\end{equation}

Here the the original event coordinates \(\left\{x, y, t\right\}\) are transformed into new coordinates \(\left\{x', y', t\right\}\). Note that the time\-stamp remains unchanged in the transformation and is omitted in Eq.~\ref{eq:transform} for simplicity. Furthermore, we assume  linear event trajectories \(\xi_{ij}\) (Eq. \ref{eq:xi_trajectory}) within the  time slice.
%\(\delta t\).

The parameters of the model denote the shift \((h_x, h_y)\) parallel to the  image plane, a motion (h_z) towards the
image plane, effectively a radial expansion of the event cloud, and a  rotation \((\theta)\) around the \(Z\) axis, effectively a circular  component of the event cloud.

\section{Camera Motion Compensation}
\label{sec_main_alg}

\begin{figure}[ht!]
\begin{center}
\includegraphics[width=0.8\columnwidth]{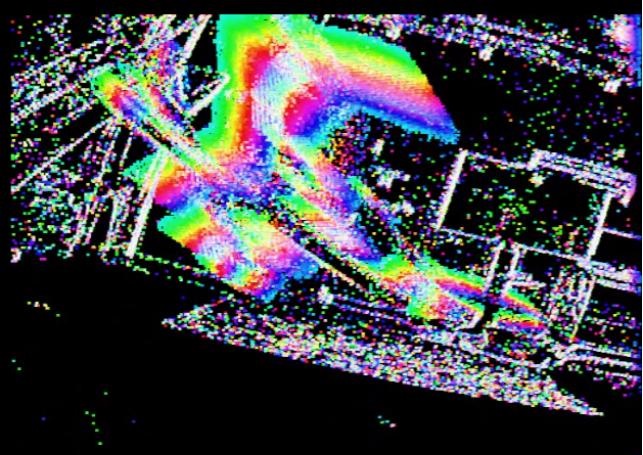}
\end{center}
\vspace{-0.0\baselineskip}
   \caption{\textit{An example output of the motion compensating algorithm - the time-image. The colors denote the average timestamps (blue is \(t_0\), green is \(t_0 + \delta t\)). On this example the separately moving object (drone) occupies the large area of the frame but the camera motion compensation still succeeds.}}
   \vspace{-1.8\baselineskip}
\label{fig:moving_large}
\end{figure}

Our pipeline (see Fig. \ref{fig:pipeline}) consists of camera motion compensation and subsequent motion inconsistency analysis to detect independently moving objects. To compensate for the global background motion, the four parameter model \(\mathcal{M}^G\) presented in Sec. \ref{sec_models} is used. The background motion is estimated, and objects are detected. Then the  background model is refined using data from the background region only (not the detected objects) (Fig. \ref{fig:moving_large}), and  four-parameter models are fit to the segmented objects for tracking. 

As outlined in Secs. \ref{sec_density} and \ref{sec_models}, $\mathcal{T}$ and $\mathcal{I}$ are local metrics of event cloud misalignment but based on different sources of data - \textit{event timestamps} and \textit{event rates}. $\mathcal{T}$ provides a poor error metric when the optimizer is close to the minima due to the event averaging scheme employed. In particular, the global error gradient functions (Eqs. \ref{eq:shift_eq} - \ref{eq:div_eq}) have very small values and are unreliable.

\begin{figure}[ht!]
\begin{center}
  \begin{minipage}[b]{0.8\linewidth}
    \includegraphics[width=1.0\textwidth]{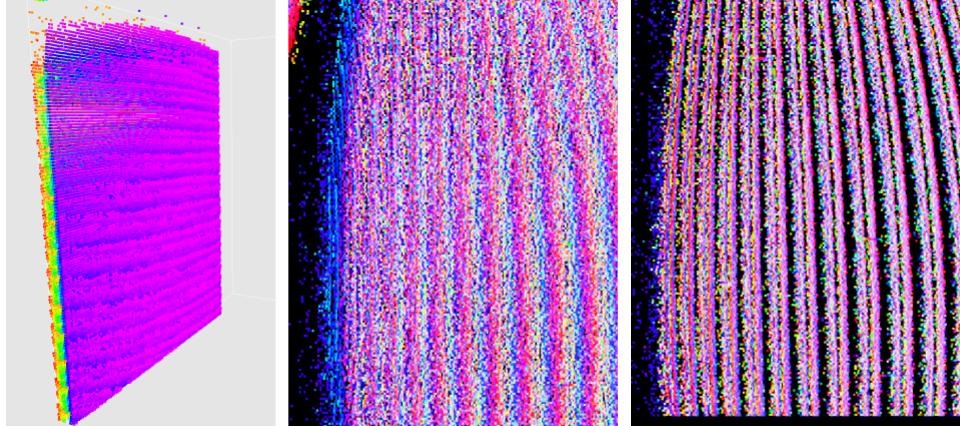}
  \end{minipage}
\end{center}
\vspace{-0.0\baselineskip}
   \caption{\small{An experiment with high texture and fast motion. The event cloud is a set of tilted planes which overlap during vertical projection. The time-image is a robust enough metric to compute the correct minimization gradient for the motion compensation. Images from left to right: 3D representation of event cloud, unminimized time-image, minimized time-image.}}
\label{fig:fast_texture}
\end{figure}

Note that, \(\mathcal{T}\) provides reliable gradients of the parameters of model \(\mathcal{M}^G\)  even in the presence of noise and fast motion, when events from different edges overlap during projection (see Fig. \ref{fig:fast_texture}).

For the aforementioned reasons, the global motion minimization is performed in two stages: coarse motion minimization on $\mathcal{T}$ and fine motion refinement on the $\mathcal{I}$.

\begin{algorithm}[H]
\caption{Global motion compensation in event space using $\mathcal{T}$.}
\label{minim_time}
\label{alg:timestamp}
\KwData{\(\mathcal{M}_{i-1}^G, C, d, \xi\)}
\KwResult{\(\mathcal{M}_{i}^G, C', \mathcal{T}\)}
\flushleft $C' \gets \text{warpEventCloud}(C, \mathcal{M}_{i-1}^G)$ \\
$\mathcal{T} \gets \text{getTimestampImage}(C', d)$ \\
$\mathcal{M}_{i}^G \gets \text{updateModel}(\mathcal{M}_{i-1}^G, \mathcal{T})$

\While{\(||{M}_{i-1}^G - {M}_{i}^G||_2 > \xi \)}{
$C' \gets \text{warpEventCloud}(C, \mathcal{M}_{i}^G)$\\
$\mathcal{T} \gets \text{getTimestampImage}(C', d)$\\
$\mathcal{M}_{i-1}^G \gets \mathcal{M}_{i}^G$\\
$\mathcal{M}_{i}^G \gets \text{updateModel}(\mathcal{M}_{i}^G, \mathcal{T})$
}
\end{algorithm}
\subsection{Coarse Global Motion Minimization on \texorpdfstring{$\mathcal{T}$}{T}}
\label{sec_alg_timestamp}
An algorithm for coarse motion compensation of the event cloud is presented in Algorithm \ref{alg:timestamp}. 

The input is the previous model \(\mathcal{M}_{i-1}^G\), original event cloud \(C\), the discretization grid size \(d\) and accuracy parameter \(\xi\). The \textit{warpEventCloud} function applies the warp field \(\psi\) as per equation (\ref{eq:transform}). The time-image \(\mathcal{T}\) is then generated on the warped event cloud \(C'\) according to (\ref{eq:ts_img}). Finally, \textit{updateModel} computes gradient images \(G_x\) and \(G_y\) (a simple Sobel operator is applied) and the gradients for the motion model parameters \(\mathcal{M}_{i}^G\) are computed with (\ref{eq:shift_eq} - \ref{eq:div_eq}). The parameters of \(\mathcal{M}_{i}^G\) are then updated in a gradient descent fashion. In this paper, the discretization parameter \(d\) has been chosen as \(0.3\) of the DVS pixel size.

\subsection{Fine Global Motion Refinement on \texorpdfstring{\(\mathcal{I}\)}{I}}
\label{sec_alg_ec}
An additional fine motion refinement is done by maximizing the density \(\mathcal{D}\) (\ref{eq:density}) of the event-count image \(\mathcal{I}\). The density \(\mathcal{D}\) function does not explicitly provide the gradient values for a given model, so variations of model parameters are used to acquire the derivatives and perform minimization. The corresponding algorithm is given in Algorithm \ref{alg:ec}.

\begin{algorithm}[H]
\caption{Global motion compensation in event space with event count image}\label{minim_count}

\KwData{\(\mathcal{M}_{i-1}^G, C, d, \xi\)}
\KwResult{\(\mathcal{M}_{i}^G, C', \mathcal{I}\)}
\flushleft $\mathcal{M}_{i}^G \gets \mathcal{M}_{i-1}^G$\\
$C' \gets \text{warpEventCloud}(C, \mathcal{M}_{i-1}^G)$\\
$\mathcal{I} \gets \text{getEventCountImage}(C', d)$\\
$\mathcal{D} \gets \text{getEventDensity}(\mathcal{I})$\\
$\mathcal{D}' \gets 0$\\

\While{\(||\mathcal{D} - \mathcal{D}'|| > \xi\)}{
$\mathcal{D} \gets \mathcal{D}'$\\
\For{Parameter p in Model \(\mathcal{M}_{i}^G\)}{
$\mathcal{D}' \gets \text{getEventDensityCloud}(C, d, \mathcal{M}_{i}^G + p)$

\If{\(\mathcal{D}' > \mathcal{D}\)}{
$\mathcal{M}_{i}^G \gets \mathcal{M}_{i}^G + p$\\
$C' \gets \text{warpEventCloud}(C, \mathcal{M}_{i}^G)$\\
$\mathcal{I} \gets \text{getEventCountImage}(C', d)$\\
}}}
\label{alg:ec}
\end{algorithm}
\section{Multiple Object Detection And Tracking}
\label{sec_track_alg}

\begin{figure}[ht!]
\begin{center}
  \begin{minipage}[b]{0.8\linewidth}
    \includegraphics[width=0.9\textwidth]{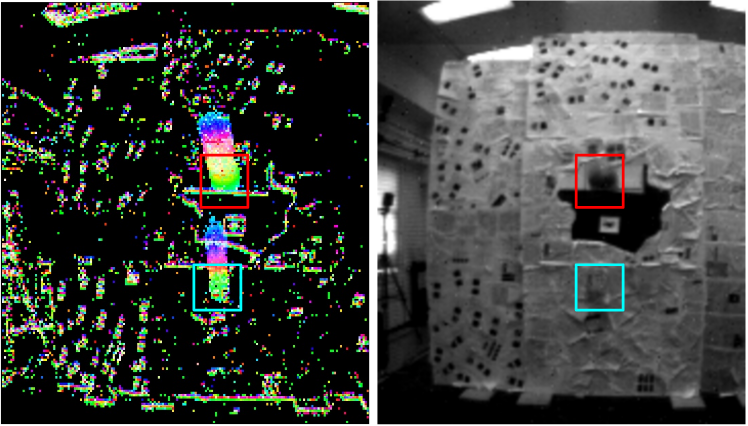}
  \end{minipage}
\end{center}
\vspace{-0.6\baselineskip}
   \caption{A frame from the 'Two Objects' dataset. The left image shows the misalignment of the two objects after the global motion compensation (green color corresponds to the most recent events). The right image is the corresponding  grayscale image - the objects (highlighted by bounding boxes) are poorly visible due to severe motion blur.}
   \vspace{-0.6\baselineskip}
\label{fig:detec_track}
\end{figure}

In this section, we describe the approach of the detection of independently moving objects by observing the inconsistencies of $\mathcal{T}$. The detected objects are then tracked using a traditional Kalman Filter. 

% Moving events are classified into independently moving objects by analyzing spatio-temporal association and model consistency constraints.

\subsection{Detection}
\label{sec_detection}
% In this section we will discuss the framework used for detecting multiple objects from a given time-image $
%\mathcal{T}$. 
We use a  simple detection scheme: We detect pixels as independently moving using a  thresholding operation and then group pixels into objects using morphological operations.
Each pixel $\left\{ i, j\right\} \in \mathcal{T}$ is associated with a score $\rho (x_i,y_j) \in [-1,1]$ defined in Eq. \ref{eq:rho}, which quantitatively denotes the misalignment of independently moving objects with respect to the background. $\rho$ is used as a measure for classifying a pixel as either background $\mathcal{B}$ or independently moving objects $\mathcal{O}_k$. 

\begin{equation}
\rho (i,j) = \frac{\mathcal{T}(i,j) -{< \mathcal{T}_{i,j} >}}{\triangle t}
\label{eq:rho}
\end{equation}
with $<  >$ denoting the mean.
Now, let us define $\mathcal{B}$ and $\mathcal{O}_k$. 

\begin{equation}
    \mathcal{B} = \{ (i,j) | \rho (i,j) \leq 0\}\qquad 
    \mathcal{O} = \{(i,j)|\rho (i,j) > \lambda \}
\end{equation}
here $\mathcal{O} = \mathcal{O}_1 \bigcup ... \mathcal{O}_n$ ($n$ is the number of independently moving objects) and $\lambda$ is a predefined minimum confidence values for  objects to be classified as  independently moving. To detect independently moving objects, we then group foreground pixels using simple morphological operations.

 % $\mathcal{B} \subset \mathcal{T}$, pixel locations can be defined as $\mathcal{B} = \{ (i,j) | \rho (i,j) \leq 0\}$ and independently moving objects, $\mathcal{O} \subset \mathcal{T}$, pixel locations can be defined as $\mathcal{O} = \{(i,j)|\rho (i,j) \textgreater \lambda \}$ where $\lamda$ is the approximation error. To determine multiple moving objects, $\{\mathcal{O}_1 ... \mathcal{O}_n \} \subset \mathcal{O}$, we employ spatial and temporal constraints. 

 % Since $\mathcal{T}$  is computed on the motion compensated event cloud \(C'\), it contains the average timestamp values. We can see that $\mathbb{E}\left(\mathcal{T}_{i,j}\right)$ is ${\textless \mathcal{T}_{i,j} \textgreater} $, where $\mathbb{E}$ is the expectation operator. Consequently, the aligned background, $\mathcal{B} \subset \mathcal{T}$, pixel locations can be defined as $\mathcal{B} = \{ (i,j) | \rho (i,j) \leq 0\}$ and independently moving objects, $\mathcal{O} \subset \mathcal{T}$, pixel locations can be defined as $\mathcal{O} = \{(i,j)|\rho (i,j) \textgreater \lambda \}$ where $\lamda$ is the approximation error. To determine multiple moving objects, $\{\mathcal{O}_1 ... \mathcal{O}_n \} \subset \mathcal{O}$, we employ spatial and temporal constraints. 

% Each independently moving object, $\mathcal{O}_{i}$, is subjected to 8-connected component (TODO) and smooth temporal gradients constraints(\ref{eq:var}). 

%\begin{equation}
% Var(\nabla(\mathcal{O}_{i}) \approx 0
% \label{eq:var}
% \end{equation}

\subsection{Tracking}
The detection algorithm presented in Subsec. \ref{sec_detection} runs in real-time (processing time $< \delta t$) and is quite  robust. To account for missing and wrong detections, especially in the presence of occlusion, we employ a simple Kalman Filter with a constant acceleration model. For the sake of brevity, we only define the state ($\mathcal{X}_{k}$) and measurement vectors ($\mathcal{Z}_{k}$) for the  $k^{\text{th}}$ object.

% Even though the update of asynchronous event cloud and detection algorithm runs in real-time, we employ a Kalman filter [cite] for smooth tracking and large occlusion handling. For compactness of this paper, we do not explain the fine details of the filter, and instead define our state and measurement matrices. The objective of the filter is to estimate states $\mathcal{X}_{1..n}$ from the measurements $\mathcal{Z}_{1..n}$, where subscript $n$ indicates the numbers of independently moving objects. For every timestamp image, $\mathcal{T}_{i}$, the filter tracks the centroid location of $\mathcal{O}_{1..n}$ as well as the model parameters of each moving objects, $\mathcal{M}_{1..n}^{G}$. The states and measurements of independently moving objects are defined as:

\begin{equation}
\begin{array}{lcl}
% \mathcal{X}_{1..n} = [\mathcal{O}_{1..n}, \mathcal{M}_{1..n}^{G} ]\\
\mathcal{X}_{k} = [\hat{x}_k,\hat{y}_k,  h_{x}, h_{y}, h_{z}, \theta, \hat{u}_k, \hat{v}_k]^T
\label{eq:state}
\end{array}
\end{equation}

\begin{equation}
\begin{array}{lcl}
% \mathcal{Z}_{1..n} = [\mathcal{O}_{1..n}, \mathcal{M}_{1..n}^{G} ]\\
\mathcal{Z}_{k} = [\hat{x}_k,\hat{y}_k, h_{x}, h_{y}, h_{z}, \theta ]^T
\label{eq:meas}
\end{array}
\end{equation}

where $\left\{\hat{x}_k, \hat{y}_k\right\}$ represent the mean coordinates of $\mathcal{O}_k$,  $h_{x}, h_{y}, h_{z}, \theta$ represent the model parameters 
%$\mathcal{M}_k^G$ 
of the object, which is obtained by motion-compensating $\mathcal{O}_k$ as described in \ref{sec_main_alg} and $\hat{u}_k, \hat{v}_k$ represents the average velocity of the $k^{\text{th}}$ object.

% TODO - conclusion (state any filter works as our detection is robust)

% \begin{algorithm}[H]
% \caption{Multiple Object Detection and Tracking}\label{detect}
% \KwData{\( \mathcal{T}_l\)}
% // Initialization when $l=0$\;
% \State $\text{initializeTrackers();}$\\
% \For{$l>0$}{
% \State $\text{getStateMeanCovariance();}$
% \State $\mathcal{O}_{1..n} \gets runDetector(\mathcal{T})$
% \State $\mathcal{M}_{1..n}^{G} \gets getModelParameters(\mathcal{O}_{1..n})$
% \State $\mathcal{Z}_{1..n} \gets getTrackerMeasurement(\mathcal{O}_{1..n}, \mathcal{M}_{1..n}^{G})$
% \State $\mathcal{X'}_{1..n}^{k} \gets trackerPredict()$
% \State $matchModels(\mathcal{Z}_{1..n}^{k}, \mathcal{X'}_{1..n}^{k})$ 
% \State $\mathcal{Z}_{1..n}^{k} \gets matchModels(\mathcal{Z}_{1..n}^{k}, \mathcal{X'}_{1..n}^{k})$ 
% \State $trackerCorrection(\mathcal{Z}_{1..n}^{k})$
%}
%
%\end{algorithmic}
%\end{algorithm}

\section{Datasets and Evaluation}

\begin{figure}[ht!]
\begin{center}
  \begin{minipage}[b]{0.8\linewidth}
    \includegraphics[width=1.0\textwidth]{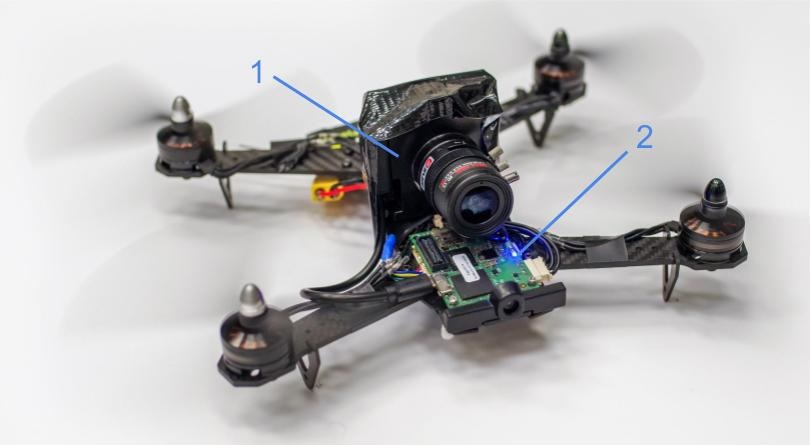}
  \end{minipage}
\end{center}
\vspace{-0.0\baselineskip}
   \caption{\small{Drone used in the dataset collection. 1 - mounted DAVIS240B camera, 2 - Customized Qualcomm Flight platform with onboard computer}}
\label{fig:e_setup}
\end{figure}

The Extreme Event Dataset (EED) used in this paper was collected using the DAVIS \cite{DAVIS_paper} sensor under two scenarios. First, it was  mounted on a quadrotor (Fig. \ref{fig:e_setup}), and second in a hand-held setup to accommodate for a variety of non-rigid camera motions. The recordings feature objects of multiple sizes moving at different speeds in a variety of lighting conditions. We emphasize the ability of our pipeline to perform detection at very high rates, and include several sequences where the tracked object changes its speed abruptly.

\begin{figure}[ht!]
\begin{center}
  \begin{minipage}[b]{0.9\linewidth}
    \includegraphics[width=1.0\textwidth]{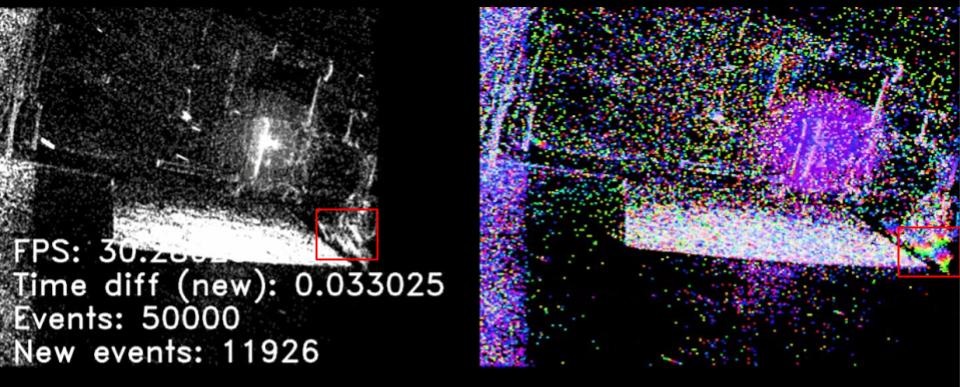}
  \end{minipage}
\end{center}
\vspace{-0.08\baselineskip}
   \caption{\small{An example from the strobe dataset. A single object moves in a dark room with a bright strobe light which produces a lot of noise on the sensor output. Since the motion model is global, the minimization and detection are tolerant to such noise. The detection output is shown with a superimposed bounding box.}}
   \vspace{-0.8\baselineskip}
\label{fig:noise}
\end{figure}

We have collected over 30 recordings in total but the centerpiece of our dataset is the 'Strobe Light' sequence. In this sequence, a periodically flashing bright light  in a  dark room creates significant noise. At the same time an object, another quadrotor is moving in the room. This is a challenge for traditional visual systems, but  our bio-inspired algorithm shows excellent performance  by leveraging the high temporal event density of the DAVIS sensor.

\subsection{Dataset collection}
\label{sec_d_collection}

The dataset was collected using the DAVIS240B bio-inspired sensor equipped with a \(3.3 mm.\) lens with a horizontal and vertical field of view of \(80^{\circ}\). Most of the sequences were created in a hand-held setting. %This simplifies the dataset collection and the creation of sequences with fast and unpredictable camera motion. 
For the quadropter sequences, we modified a Qualcomm Flight$^{\text{TM}}$ \cite{SnFlight} platform to connect the DAVIS240B sensor to the onboard computer and collect data in a realistic scenario. 

The setup of the quadrotor+sensor platform can be seen in Fig. \ref{fig:e_setup}. The overall weight of the fully loaded platform is \(\approx 500 g.\) and it is equipped with the Snapdragon APQ8074 ARM CPU, featuring 4 cores with up to 2.3GHz frequencies.

\subsection{Computation Times}
On a single thread of Intel Core$^{\text{TM}}$ i7 3.2GHz processor, Algorithms \ref{alg:timestamp} and \ref{alg:ec} take $10ms$ and $7ms$ on  average  for a single iteration step. However, both  algorithms are based on a warp-and-project \(\Pi\{\phi(C)\}\) operation which is highly parallelizable and thus well fit for  implementation on a Graphical Processing Unit (GPU) or a Field-Programmable Gate Array (FPGA) to acquire very low latency and high processing speeds.

While a low level hardware implementation is beyond the scope of this paper, we tested a prototype of the algorithm on an NVIDIA Titan X Pascal$^{\text{TM}}$ GPU with CUDA acceleration. A single iteration for Algorithms \ref{alg:timestamp} and \ref{alg:ec} takes $0.01ms$ and $0.003ms$ on  average, respectively, which is a 1000X and 2333X speed-up. We have empirically found that the minimization converges on average in less than 30 iterations which ensures a faster-than-real-time computation speed with a high margin.

% CUDA and have achieved the processing speeds of \(0.01 ms.\) for Algorithm \ref{alg:timestamp} and \(0.003 ms.\) for Algorithm \ref{alg:ec} per minimization iteration on a . We have established that the minimization converges on average in less than 30 iterations which ensures a faster-than-real-time computation speeds with a high margin.

%\textbf{Note.} \textit{Computational performance considerations - both algorithms (Algorithm \ref{alg:timestamp} and Algorithm \ref{alg:ec}) are based on a warp-and-project \(\Pi\{\phi(C)\}\) operation which is highly parallelizable and is easy to implement on the Graphical Processing Unit (GPU) or a Field-Programmable Gate Array (FPGA) to acquire very low latency and high processing speeds. While the low level hardware implementation is beyond the scope of this paper, we have created a test implementation of the algorithms using Nvidia CUDA and have achieved the processing speeds of \(0.01 ms.\) for Algorithm \ref{alg:timestamp} and \(0.003 ms.\) for Algorithm \ref{alg:ec} per minimization iteration on a Titan X Pascal GPU. We have established that the minimization converges on average in less than 30 iterations which ensures a faster-than-real-time computation speeds with a high margin.
%}

\begin{figure*}[h]
\begin{center}
\resizebox{\linewidth}{!}{\begin{tabular}{cccc}  
   \includegraphics[width=0.24\textwidth]{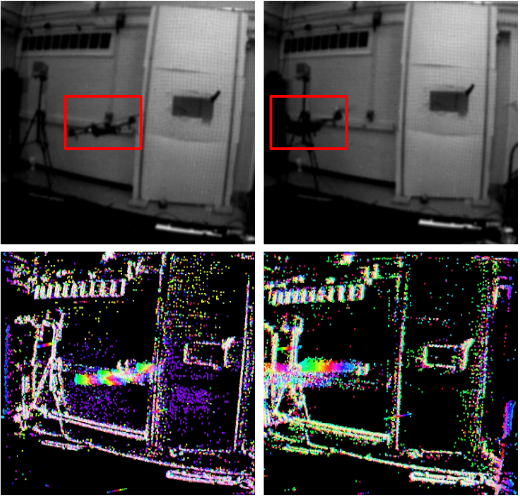}&
   \includegraphics[width=0.24\textwidth]{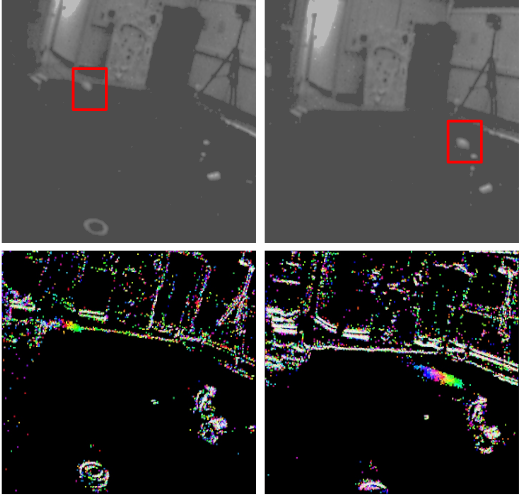}&
   \includegraphics[width=0.24\textwidth]{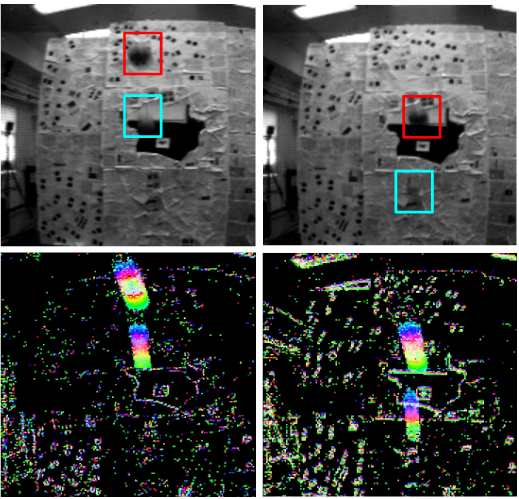}&
   \includegraphics[width=0.24\textwidth]{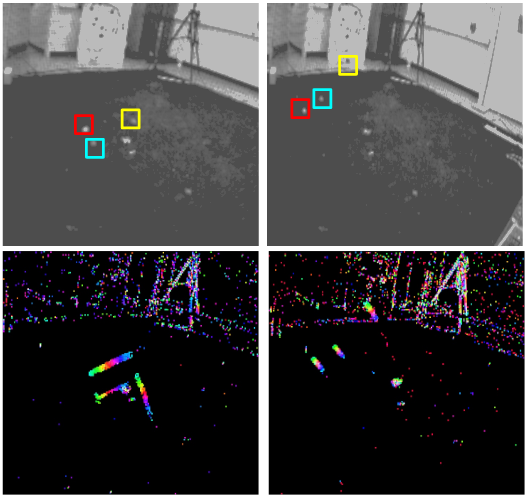}\\
   (a) & (b) & (c) & (d)\\
   \includegraphics[width=0.24\textwidth]{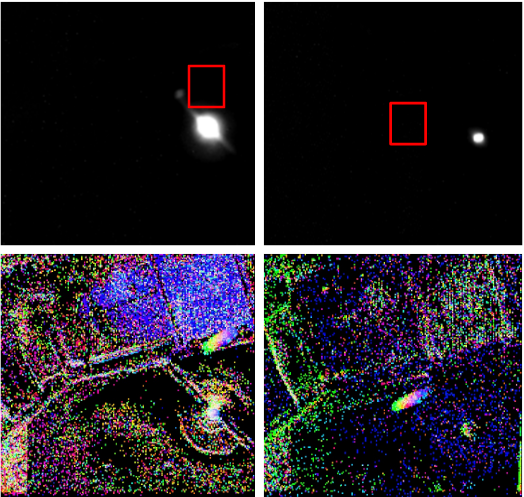}&
   \includegraphics[width=0.24\textwidth]{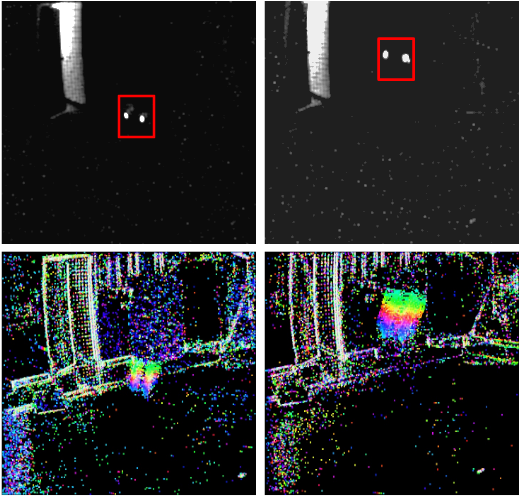}&
   \includegraphics[width=0.24\textwidth]{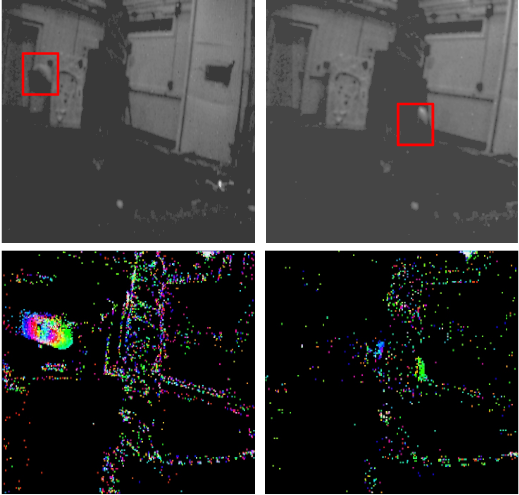}&
   \includegraphics[width=0.24\textwidth]{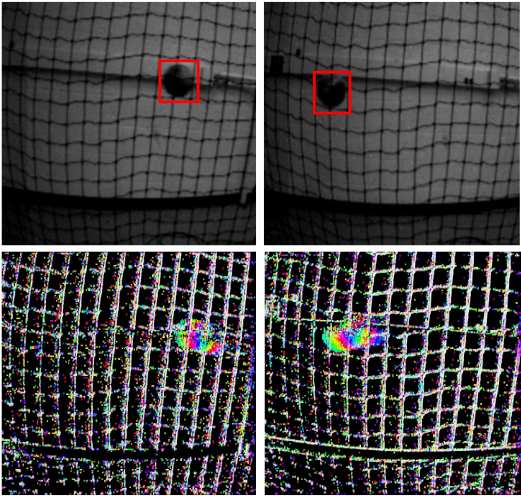}\\
   (e) & (f) & (g) & (h)\\
\end{tabular}}  
\vspace{-1.3\baselineskip}
\end{center}
   \caption{\small{Sample frames from the EED dataset are shown. Top row shows RGB frames with superimposed bounding boxes for tracked object. The bottom row displays the corresponding time-images. (a) - Remote Controlled Quadrotor recording, (b), (c) and (d) - Single, two and three simultaneously moving objects, (e) - Strobe Light sequence, (f) - low lighting (no strobe), (g) - An ''Occluded Sequence'', (h) - ''What is a Background?'' sequence}}
\label{fig:dataset}
\end{figure*}

The recordings are organized into several sequences according to the nature of scenarios present in the scenes. All recording feature a variety of camera motions, with both rotational and translational motion (See fig. \ref{fig:dataset}):

\begin{itemize}
\item \textit{''Fast Moving Drone''} - A sequence featuring a single small remotely controlled quadrotor. Quadrotor has a rich texture and moves across various backgrounds in daylight lighting conditions following a variety of trajectories.

\item \textit{''Multiple objects ''} - This sequence consists of multiple recordings with 1 to 3 moving objects under normal lighting conditions. The objects are simple, some of them have little to no texture. The objects move at a variety of speeds, either along  linear trajectories or they bump from  a surface.

\item \textit{''Lighting variation''} - A strobe light flashing at frequencies of 1-2 Hz  %with periods from \(0.5 s.\) to \(1.0 s.\) 
was placed in a dark room to produce a lot of noise in the event sensor. This is an extremely challenging sequence, otherwise similar to the \textit{''Fast Moving Drone''}.

\item \textit{''What is a Background?''} - In most tracking algorithm evaluations, object moves in front of the background. The following toy sequence was included, to show that it is possible to track an  object even when the background occupies the space in between the camera and the object: A simple object was placed behind the net and the motion could only be seen through the net. Recordings contain a variety of distances between the net and the camera and the object is thrown at different speeds.

\item \textit{''Occluded Sequence''} - The sole purpose of this sequence is to test the reliability of tracking in scenarios when detection is not possible for a small period of time. Several recordings feature object motion in occluded scenes.
\end{itemize}

\subsection{Metrics and Evaluation}
\label{sec_big_picture}

\begin{figure}[ht!]
\begin{center}
  \begin{minipage}[b]{0.8\linewidth}
    \includegraphics[width=1.0\textwidth]{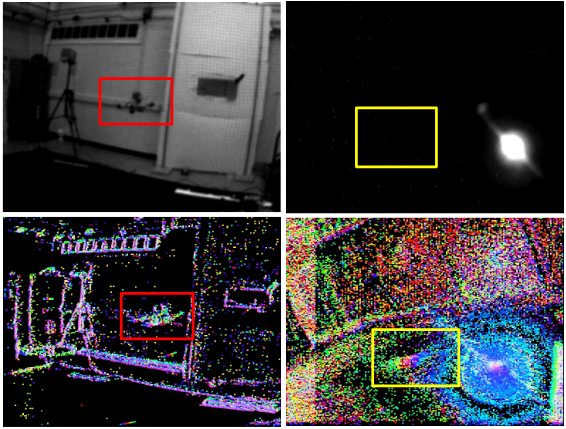}
  \end{minipage}
\end{center}
\vspace{-0.0\baselineskip}
   \caption{\small{Common failure cases. Top row: RGB frames, bottom row: time-images. Left: A setting featuring an object which does not move with respect to the background - a failure case for the detection stage, even though the edges are visible. Right: High noise scenario with the tracked object not visible in the gray-scale image.}}
\label{fig:failure_0}
\end{figure}

\begin{table*}[t]
\caption{\small{Evaluation of the pipeline on the proposed dataset}}
\begin{center}
\resizebox{1\linewidth}{!}{\begin{tabular}{|l|c|c|c|c|c|}
\hline
Sequence & ''Fast Moving Drone'' & ''Multiple objects '' & ''Lighting variation'' & ''What is a Background?'' & ''Occluded Sequence'' \\
\hline\hline
\textbf{Success Rate} & \textbf{92.78\%} & \textbf{87.32\%} & \textbf{84.52\%} & \textbf{89.21\%} & \textbf{90.83\%} \\
\hline
\end{tabular}}
\vspace{-0.0\baselineskip}
\end{center}
\label{table:q_results}
\end{table*}

We define our evaluation metrics in form of success rate: We have acquired the ground truth by hand labeling the RGB frames of the  recordings. We then computed  a separate success rate for every time slice corresponding to an RGB frame from the DAVIS sensor as the percentage of the detected objects with at least \(50\%\) overlap with the object visible in the RGB frame. The mean of those scores for all sequences is reported in Table \ref{table:q_results}.

Although we did not discover sequences where the motion-compensation pipeline performs poorly, the particular difficulty for the tracking algorithm were the strobe light scenes where noise from the strobe light completely covered the tracked object and prevented detection - the noise on such scenes was even further amplified by the low lighting conditions.

Interestingly, a high performance was achieved on the \textit{''What is a Background?''} sequence. The object was partially occluded by the net (located between the camera and the object), but in favor for the algorithm, the high texture of the net allowed for a robust camera motion compensation.

other challenging time sequences were the ones which featured objects whose paths crossed in the image. The Kalman filter often was able to distinguish between the tracked objects based on the difference in states (effectively, the difference in the previous motion)

To conclude this section, we feel the need to discuss some common failure cases.  Figure \ref{fig:failure_0} demonstrates a frame from the ''Fast Moving Drone'' sequence - the motion of the drone with respect to the background is close to zero. The motion-compensation stage successfully compensates the camera motion but fails to recognize a separately moving object at that specific moment of time.

Another cause for failure is the presence of severe noise, as demonstrated by the ''Lighting variation'' sequence. The motion compensation pipeline is robust only if a sufficient amount of  background is visible. Figure \ref{fig:failure_0} (right image) demonstrates that in some conditions too much noise is projected on the time image which renders both motion compensation and detection stages unreliable.

\section{Conclusions}
We argue that event-based sensing can fill a void in the area of robotic visual navigation. Classical frame-based Computer and Robot vision has great challenges in scenes with fast motion, low-lighting, or changing lighting  conditions. We believe that event-based sensing coupled with active purposive algorithmic approaches can provide the necessary solutions.
Along this thinking, in this paper we  have presented the first  event-based only method for motion segmentation under unconstrained conditions (full 3D unknown motion and unknown scene). The essence of the algorithm lies in a method for efficiently and robustly  estimating the  effects of 3D camera motion from the event stream. Experiments in challenging conditions of fast motion with multiple moving objects and lighting variations demonstrate the usefulness of the method. 

Future work will extend the method to include more elaborate clustering and segmentation. The goal is  to implement the 3D motion estimation and clustering in a complete iterative approach to accurately estimate 3D motion while detecting all moving objects, even those that move similar to the camera.

\section{Acknowledgements}
The  support of the National Science Foundation under awards  SMA 1540917 and  CNS 1544797,  a  UMD-Northrop Grumman seed grant, and  ONR under award  N00014-17-1-2622 are  gratefully acknowledged.

%{\small
%\bibliography{egbib}
%\bibliographystyle{IEEEtran}
%}

\bibliographystyle{IEEEtranS}
\bibliography{egbib}

\end{document}